%% file: main.tex
\documentclass[10pt,twocolumn,letterpaper]{article} 

\usepackage{avss}
\usepackage{times}
\usepackage{epsfig}
\usepackage{graphicx}
\usepackage{amsmath}
\usepackage{amssymb}
\usepackage{multirow}
\usepackage{pifont}
\usepackage{hyperref} 

\avssfinalcopy 

\def\avssPaperID{111} 

\ifavssfinal\fi
\begin{document}

\title{3D-PointZshotS: Geometry-Aware 3D Point Cloud Zero-Shot Semantic Segmentation Narrowing the Visual-Semantic Gap \vspace{-0.4cm}}

\author{Minmin Yang \and Huantao Ren \and Senem Velipasalar\\
Electrical Engineering and Computer Science Dept., Syracuse University\\
Syracuse, NY, USA\\
{\tt\small myang47@syr.edu } 
{\tt\small hren11@syr.edu } 
{\tt\small svelipas@syr.edu }
}
\vspace{-0.4cm}
\maketitle

\begin{abstract}
   Existing zero-shot 3D point cloud segmentation methods often struggle with limited transferability from seen classes to unseen classes and from semantic to visual space. To alleviate this, we introduce 3D-PointZshotS, a geometry-aware zero-shot segmentation framework that enhances both feature generation and alignment using latent geometric prototypes (LGPs). Specifically, we integrate LGPs into a generator via a cross-attention mechanism, enriching semantic features with fine-grained geometric details. To further enhance stability and generalization, we introduce a self-consistency loss, which enforces feature robustness against point-wise perturbations. Additionally, we re-represent visual and semantic features in a shared space, bridging the semantic-visual gap and facilitating knowledge transfer to unseen classes.~Experiments on three real-world datasets, namely ScanNet, SemanticKITTI, and S3DIS, demonstrate that our method achieves superior performance over four baselines in terms of harmonic mIoU. The code is available at \href{https://github.com/LexieYang/3D-PointZshotS}{Github}.
  \vspace{-0.4cm} 
\end{abstract}

\input{sec/1_intro}

\input{sec/2_related}
\input{sec/3_method}

\input{sec/4_experiments}

\input{sec/5_conclusion}

{\small
\bibliographystyle{ieee}
\bibliography{egbib}
}

\input{sec/suppl}

\end{document}

%% file: sec/1_intro.tex
\vspace{-0.3cm}
\section{Introduction}
\label{sec:intro}

Two essential aspects of achieving successful zero-shot segmentation (ZSS) in 3D point clouds are effective alignment between semantic and visual features and good generalization from seen to unseen classes.~A generative framework was proposed in \cite{michele2021generative} to generate fake unseen object features by leveraging class-level semantic information.~Yet, synthetic features lack geometric diversity and consistency, since the generator is merely given the high-level semantic features.~Moreover, as the generator is only trained on seen class samples, this may cause limited transferability and low generalization to unseen classes.~Following the generative framework~\cite{michele2021generative},~\cite{yang2023zero} introduced three modules to improve the generated visual features and transferability from semantic to visual space. However, no matter how diverse the generated visual features are, the domain gap between the semantic and visual features still exists. Moreover, there is no module for explicitly transferring and adapting the knowledge from seen classes to unseen classes. 

To address the aforementioned issues, we explore the role of the 3D geometric structure of point clouds by utilizing it to bridge the gap between visual and semantic spaces, and enabling the capture and transfer of shared geometric patterns across seen and unseen classes.
For example,~\cite{xu2023generalized} addresses the generalized few-shot point cloud segmentation problem by introducing ``geometric words", which encode common geometric components across base and novel classes.
However, these geometric words are pre-extracted and remain fixed during training, limiting their adaptability. In contrast, ~\cite{chen2023bridging} proposes learnable geometric primitives for visual feature representation. Both geometric words~\cite{xu2023generalized} and geometric primitives~\cite{chen2023bridging} have been shown to represent fundamental geometric structures, such as cones and cylinders.~Building on these ideas, we introduce Latent Geometric Prototypes (LGPs), which are \emph{learned} representations that dynamically capture spatial relationships and local structures.~Unlike predefined geometric structures, LGPs adaptively model geometric features, enabling more effective alignment between visual and semantic spaces for ZSS.

Expanding on the concept of LGPs as a bridge between semantic and visual spaces, we present 3D-PointZshotS, which is a geometric consistency-aware zero-shot segmentation framework.~Our proposed framework incorporates LGPs in both feature generation and alignment through two key strategies:~(i) \textbf{Geometric Consistency-Aware Generator}, where LGPs are integrated into the generator to enhance geometry-awareness and ensure that synthesized features capture fine-grained geometric structures.~We employ an InfoNCE-based self-consistency loss to enforce robustness against variations and perturbations;~(ii) \textbf{Cohesive Visual-Semantic Representation Space via LGPs}: we re-represent both visual and semantic features using a shared set of LGPs, creating a unified representation space that aligns abstract semantic concepts with detailed visual features.~By sharing LGPs across classes, we leverage transferable geometric knowledge, allowing the model to effectively capture common geometric structures present in both seen and unseen categories. To further enhance alignment between the visual and semantic embeddings, we employ InfoNCE loss, which encourages closer alignment between matching class embeddings while ensuring separation between different classes.

In summary, we introduce the 3D-PointZshotS framework that utilizes a set of latent geometric prototypes in both feature generation and alignment. The main contributions of this work include the following: \textbf{(1)} We design a geometric consistency-aware generator with InfoNCE-based self-consistency loss, helping synthesize robust and consistent features and enhancing generalization to unseen classes; \textbf{(2)} We propose a shared representation space by re-representing both visual and semantic features in terms of similarities to LGPs, bridging the gap between abstract semantics and detailed visual features, \textbf{(3)} We conduct experiments on S3DIS, ScanNet, and SemanticKITTI datasets demonstrating the superiority of the proposed method, \textbf{(4)} We perform ablation studies to show the contributions of various components of our framework.

%% file: sec/2_related.tex
\section{Related Work}
\label{sec:related}
Leveraging CLIP’s \cite{radford2021learning} robust zero-shot performance, researchers~\cite{liu2024affinity3d,peng2023openscene} have transferred knowledge from well-trained 2D vision-language models to point clouds, enabling accurate zero-shot point cloud segmentation. \cite{peng2023openscene} proposed a framework that aligns 3D points with vision-language embeddings, enabling flexible queries beyond traditional category labels. Similarly, \cite{liu2024affinity3d} refined CLIP-generated pseudo labels through instance-level semantic affinity and visibility estimation.
However, these zero-shot segmentation approaches rely on the simultaneous availability of 2D RGB images and 3D point cloud data, imposing higher demands on data acquisition. Other works~\cite{chen2023bridging,michele2021generative,yang2023zero} focus on utilizing semantic embeddings alone to assist zero-shot point cloud segmentation, which aligns with our approach. \cite{chen2023bridging} proposed a transductive ZSS model by learning shared geometric primitives between the seen and unseen classes and then performing fine-grained alignment of the language semantics with the learned geometric primitives. The comparable methods to our work are~\cite{michele2021generative,yang2023zero}, since they do not use training samples from unseen classes either and solely employ semantic features as auxiliary modalities. They generate synthetic features for unseen classes using semantic information to train the classifier for zero-shot transfer. However, since the generator is merely based on high-level semantic features, the resulting features exhibit limited variety in geometry. 

Therefore, in this work, we incorporate latent geometric prototypes (LGPs) in the generator and optimize it with the self-consistency loss, making it better synthesize visual features based on shared geometric characteristics rather than relying solely on class-specific semantic information. The LGPs are also incorporated in the semantic-visual feature alignment step to bridge the semantic-visual domain gap.


%% file: sec/3_method.tex
\section{Proposed Method}
\label{sec:method}

\subsection{Task Formulation}
\vspace{-0.1cm}
For zero-shot point cloud semantic segmentation, we have a visual space $\mathcal{V}$ to represent the visual features of point clouds, and a semantic space $\mathcal{S}$ to represent the semantic features of categories. Each space is divided into two \textit{non-overlapping} subspaces corresponding to seen and unseen classes, denoted by $\mathcal{V}^{s}$ and $\mathcal{V}^u$ for the visual space, and $\mathcal{S}^s$ and $\mathcal{S}^u$ for the semantic space, respectively. The sets of seen and unseen classes are denoted by $C^s$ and $C^u$. Let's define a dataset $\mathcal{D}$ that consists of a point cloud set $\mathcal{P}$, ground-truth label set $\mathcal{Y}=\{\mathcal{Y}^s, \mathcal{Y}^u\}$ and class-wise semantic embeddings $\mathcal{T}=\{\mathcal{T}^s,\mathcal{T}^u\}$.
Since our approach aims for the challenging \textit{inductive generalized} ZSS, the training set $\mathcal{D}_{train}$ can be denoted as $\mathcal{D}_{train}=\{(p_i,y_i,t_i)\ |\ \forall i, y_i \in C^s\}$, and testing set can be denoted as $\mathcal{D}_{test}=\{(p_i,y_i,t_i)\ |\ \forall i, y_i \in C^s \cup C^u\}$. 

\subsection{Model Structure}
\begin{figure*}[t]
  \centering
   \includegraphics[width=0.8\linewidth]{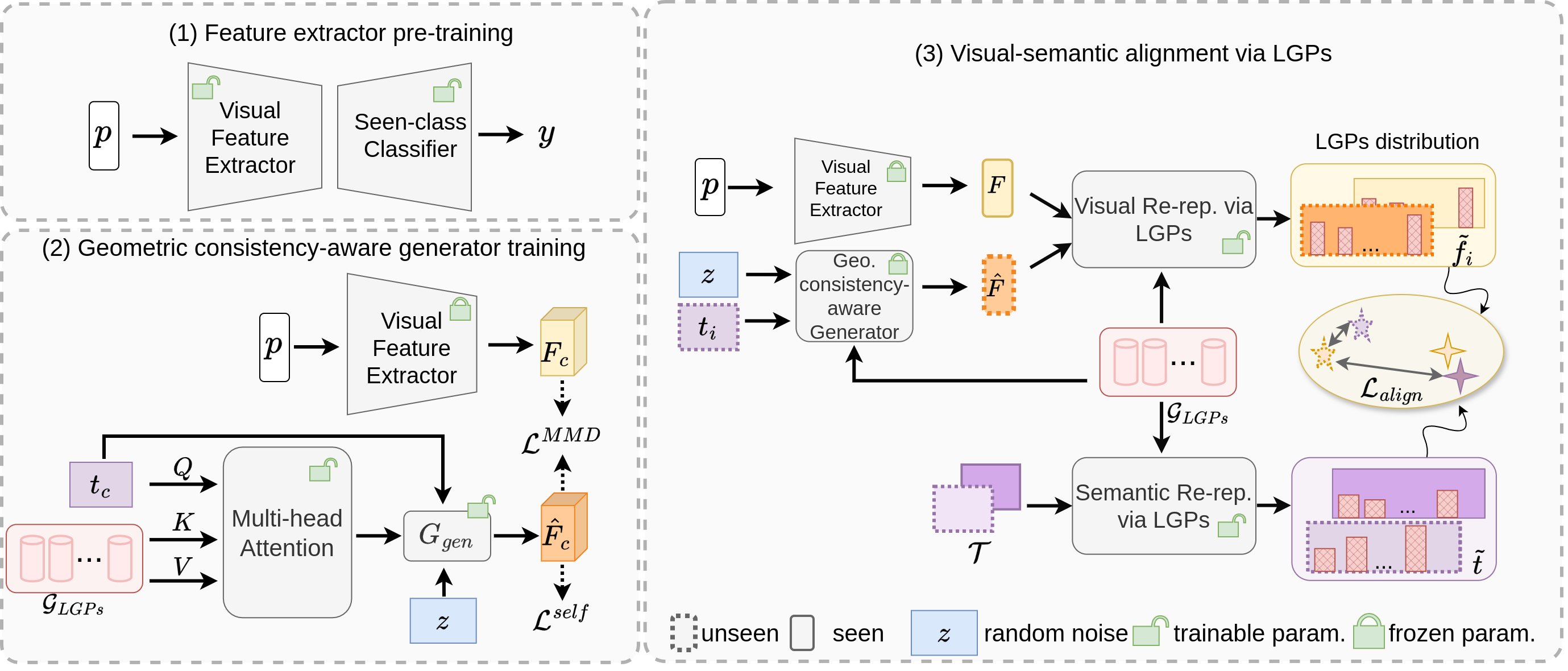}

   \caption{Three-step training procedure: (1) Feature extractor pre-training on seen classes, (2) Geometric consistency-aware generator training on seen class data. $Q$, $K$, and $V$ represent queries, keys and values, respectively. (3) Visual-semantic alignment via LGPs.}
   \label{fig:train}
   \vspace{-0.4cm}

\end{figure*}

\vspace{-0.1cm}
As shown in Fig.~\ref{fig:train}, the training pipeline for our framework consists of three main steps, which are designed to generate robust visual features and align visual and semantic features effectively by leveraging LGPs: \textbf{(1) Pre-training:} The visual feature extractor and classifier are pre-trained using data from seen classes to ensure reliable visual feature extraction. \textbf{(2) Generator training:} The visual feature extractor is frozen and the geometric consistency-aware generator is trained on seen class data by incorporating LGPs and semantic features, and optimized with self-consistency loss and maximum mean discrepancy loss. This step focuses on synthesizing robust visual features in a way that incorporates geometric understanding. \textbf{(3) Visual-semantic alignment via LGPs:}~With the visual feature extractor and generator frozen, LGPs are trained alongside both the Visual-LGPs and Semantic-LGPs re-representation modules to achieve better visual-semantic alignment.

\subsection{Latent Geometric Prototypes (LGPs)}
\vspace{-0.1cm}
Point clouds inherently encapsulate geometric attributes, composed of fundamental structures, such as cones, pyramids, and cuboids.~By combining these geometric primitives
\cite{chen2023bridging}, each class can express its inherent spatial characteristics, 
capturing the fine-grained  details of 3D objects.~Moreover, these geometric primitives provide transferable knowledge, since both seen and unseen classes often share local geometric structures. However, the expressiveness of these geometric primitives~\cite{chen2023bridging} is inherently constrained due to their limited diversity and fixed number, restricting their ability to fully represent complex 3D scenes. To address this limitation, 
we introduce a set of $M$ \textit{latent geometric prototypes (LGPs)}, denoted as $\mathcal{G}_{LGPs}=\{g_i\}^M_{i=1} \in \mathbb{R}^{M \times d}$, which reside in high-dimensional feature space. In Sec.~\ref{subsec:generator} and~\ref{subsec:vs_align}, we illustrate how these LGPs fulfill dual roles in feature generation and alignment, respectively. 

\subsection{Geometric Consistency-Aware Generator}
\label{subsec:generator}
\vspace{-0.1cm}
While 3D point clouds encapsulate fine-grained geometric details, semantic textual representations often provide high-level, abstract descriptions.~Thus, generating point embeddings based solely on semantic embeddings is challenging, since it requires bridging the gap between the symbolic abstraction of language and the nuanced geometric structures within 3D data.~Although~\cite{he2023primitive} addresses a related issue by learning image-based primitives (e.g., color, shape, and texture) to enrich image generation, our approach diverges by introducing learnable geometric prototypes specifically designed for 3D data. Unlike the image-based primitives,
our LGPs capture the intrinsic geometric attributes of point clouds.
To generate $N_c$-many points for class $c$, we first replicate the semantic representation $N_c$ times ($\Tilde{t}_c \in \mathbb{R}^{N_c\times d_t}$) and then project $\Tilde{t}_c$ into the same $d$-dimensional space as the LGPs using a linear layer, resulting in $\hat{t}_c \in \mathbb{R}^{N_c\times d}$.~We then apply a cross-attention module to $\hat{t}_c$ and the LGPs. 
This results in a geometry-aware semantic representation $\hat{t}'_c$, enriching the semantic features of class $c$ with fine-grained geometric information. Finally, we combine the transformed semantic representation $\hat{t}_c$, the geometry-aware semantic representation $\hat{t}'_c$, and random noise $z \in \mathbb{R}^d$ through summation, and feed the resulting representation into the generator $G_{gen}$ to synthesize the visual feature $\hat{F}_c=G_{gen}( \hat{t}'_c +  \hat{t}_c + z)$.
We apply Maximum Mean Discrepancy (MMD)~\cite{li2015generative} to ensure that the generated features match the distribution of real features.~The MMD loss is computed as: \vspace{-0.1cm}
\begin{equation}
\scriptsize
\mathcal{L}^{MMD}_c =\!\!\! \sum_{f,f' \in F_c} \!\!k(f,f') + \!\!\!\sum_{\hat{f},\hat{f}' \in \hat{F}_c} k(\hat{f},\hat{f}') - 2 \sum_{f \in F_c} \sum_{\hat{f} \in \hat{F}_c} k(f, \hat{f}),
\end{equation}
where $k(\cdot,\cdot)$ is the Gaussian kernel function, $f$ and $f'$ ($\hat{f}$ and $\hat{f}'$) are features sampled from $F_c$ ($\hat{F}_c$).

What makes our generator different from~\cite{he2023primitive} is that we introduce an InfoNCE-based self-consistency loss to enhance the robustness of the synthesized features. Specifically, we randomly sample $N_k$ point features twice from $\hat{F}_c \in \mathbb{R}^{N_c \times d}$, obtaining two sets of point features, $\hat{F}'_c$ and $\hat{F}''_c \in \mathbb{R}^{N_k \times d}$.~This process enforces the model to preserve feature consistency across different augmentations of the same point cloud.~We set the $\hat{F}'_c$ and $\hat{F}''_c$ as the positive pair, and randomly sample real features from other classes as negative samples to calculate the self-consistency loss $
\footnotesize 
    \mathcal{L}_c^{self} = -log \frac{exp({D(\hat{F}'_c,\hat{F}''_c)/\tau_1})}{exp({D(\hat{F}'_c,\hat{F}''_c)/\tau_1}) + \sum\limits_{i \in \mathcal{Y}^s, i \neq c} exp({D(\hat{F}'_c,F_i)/\tau_1})},
$
which promotes self-consistency and inter-class differentiation, thereby producing more robust and distinguishable feature representations. The final loss function $
    \mathcal{L}_G = \sum\limits_{c \in C^s} \mathcal{L}_c^{MMD} + \lambda_1 \times \mathcal{L}_c^{self}, $
where $\lambda_1$ is the loss weight.

\subsection{Visual-Semantic Alignment via LGPs}
\label{subsec:vs_align}
\vspace{-0.1cm}
Directly aligning visual and semantic features is challenging, since 
visual features capture spatial and geometric structures, while semantic features encode abstract categorical meanings. This discrepancy leads to misalignment, making it difficult to establish a direct correspondence between two representations. Inspired by \cite{chen2023bridging}, we transform both visual and semantic features into a common space through LGPs to bridge this gap. While \cite{chen2023bridging} only re-represents visual features
 to transfer knowledge learned from seen classes to unseen classes, we re-represent both point and semantic features through our LGPs and impose a shared geometric structure that enhances compatibility between the two modalities. The dual transformations ensure that visual features retain semantic relevance while semantic features gain geometric awareness, ultimately facilitating a more structured and effective alignment.

\textbf{Visual Re-representation via LGPs.}
First, we re-represent the point feature $\hat{f}_{i}$ by calculating its similarity to each latent geometric prototype, allowing each point's feature to be represented as a weighted combination of LGPs. The point feature $\Tilde{f}_{i} = [w_{i,1}^p, w_{i,2}^p, ..., w_{i,M}^p]$, where
$\small w_{i,j}^p = Softmax (\frac{\psi(\hat{f}_{i}) \cdot \phi(g_j)^T}{\sqrt{d}}),$
with $\psi(\cdot)$ and $\phi(\cdot)$ being the linear layers. $\Tilde{f}_{i}$ can be interpreted as a probability distribution since $\sum_{j=1}^{j=M} w_{i,j}^p = 1$.

The fundamental reason for probability-based representation working is that it captures the composition and importance of different geometric prototypes, similar to how a bag-of-words model captures the meaning of a sentence through word probability.~Moreover, it improves the model's discriminative power by enabling it to differentiate between classes that may have similar overall shapes but differ in their fine geometric details. 

\textbf{Semantic Re-representation via LGPs.}
Second, to construct a coherent representation for both 3D points and semantics, we re-represent the semantic representation by calculating their similarity to each LGP as well.~Thus, the semantic representation for class $c$ is expressed as $ \Tilde{t}_{c} = [w_{c,1}^t, w_{c,2}^t, ..., w_{c,M}^t],$
where $\scriptsize  w_{c,j}^t = softmax (\frac{\sigma(\hat{t}_{c}) \cdot \theta(g_j)^T}{\sqrt{d}}),$ and $\sigma(\cdot)$ and $\theta(\cdot)$ are linear layers.~By expressing both the semantic and visual representations in terms of similarities to shared LGPs, we create a common feature space. This shared representation bridges the gap between the abstract semantics and the detailed, geometry-oriented visual features.

\textbf{Visual and Semantic Alignment in Geometric Space.} Lastly, with both semantic and visual features now represented as distributions over LGPs, we apply InfoNCE loss to align representations from two modalities. When semantic and visual embeddings correspond to the same class, the InfoNCE loss pulls them closer, whereas embeddings from different classes are pushed apart, ensuring better feature separation and alignment. The loss is:
$
\scriptsize
    \mathcal{L}_{align} = -log \sum\limits_{c \in C^s \cup C^u}\sum\limits_{i=1}^{N_c} \frac{e^{D(\Tilde{t}_c,\Tilde{f}_{i})/ \tau_2}}{\sum\limits_{k \in C^s \cup C^u} e^{D(\Tilde{t}_k,\Tilde{f}_{i})/\tau_2}},
$
where $\tau_2$ is the temperature term.
\vspace{-0.15cm}
\subsection{Inference}
\vspace{-0.1cm}
During inference, we use the extracted point features $F = \{f_i\}_{i=1}^{N}$ from the scene-level point cloud data, and re-represent the point features as $\Tilde{F} = \{\Tilde{f_i}\}_{i=1}^{N}$ and semantic features as $\Tilde{T}=\{\Tilde{t}_i\}_{i=1}^{N^s+N^u}$.~We predict the label $c^*_i$ for the $i^{th}$ point cloud as: \vspace{-0.2cm}
\begin{equation}
\footnotesize 
    c^*_i = \arg \underset{c}{\max} \; \frac{exp(D(\Tilde{f}_i, \Tilde{t}_c)/\tau_2)}{\sum\limits_{\hat{c} \in C^s \cup C^u} exp(D(\Tilde{f}_i, \Tilde{t}_{\hat{c}})/\tau_2)}.
\end{equation}
\vspace{0.1cm}

%% file: sec/4_experiments.tex
\vspace{-0.25cm}
\section{Experiments}
\label{sec:exp}

\begin{table*} 
\centering
\resizebox{\textwidth}{!}{
\begin{tabular}{l|cc|cccc|cccc|cccc}
\hline
& \multicolumn{2}{c|}{Training set} & \multicolumn{4}{c|}{S3DIS} & \multicolumn{4}{c|}{SemanticKITTI} & \multicolumn{4}{c}{ScanNet}\\
& \multirow{2}{*}{backbone} & \multirow{2}{*}{segmentor} & \multicolumn{3}{c|}{mIoU} & \multirow{2}{*}{HmIoU} & \multicolumn{3}{c|}{mIoU} & \multirow{2}{*}{HmIoU}  & \multicolumn{3}{c|}{mIoU} & \multirow{2}{*}{HmIoU} \\
& & &  $C^S$ & $C^U$ & \multicolumn{1}{c|}{All} & & $C^S$ & $C^U$ & \multicolumn{1}{c|}{All} & & $C^S$ & $C^U$ & \multicolumn{1}{c|}{All} & \\
\hline
\multicolumn{15}{l}{\textit{\textbf{Supervised methods with different levels of supervision}}}  \\
Full supervision & $C^S \cup C^U$ & $C^S \cup C^U$  & 74.0 & 50.0 & 66.6 & 59.6 & 59.4 & 50.3 & 57.5 & 54.5 & 43.3 & 51.9 & 45.1 & 47.2 \\
ZSL backbone & $C^S$ & $C^S \cup C^U$ & 60.9 & 21.5 & 48.7 & 31.8 & 52.9 & 13.2 & 42.3 & 21.2 & 41.5 & 39.2 & 40.3 & 40.3 \\
ZSL-trivial & $C^S$ & $C^S$ & 70.2 & 0.0 & 48.6  & 0.0 & 55.8 & 0.0 & 44.0 & 0.0 & 39.2 & 0.0 & 31.3 & 0.0 \\
\hline
\hline
\multicolumn{15}{l}{\textit{\textbf{Generalized zero-shot-learning methods}}} \\
ZSLPC-Seg~\cite{cheraghian2019zero} & $C^S$ & $C^U$ & 5.2 & 1.3 & 4.0 & 2.1 & 26.4 & 10.2 & 21.8 & 14.7 & 16.4 & 4.2 & 13.9 & 6.7 \\
DeviSe-3DSeg~\cite{frome2013devise} & $C^S$ & $C^U$ & 3.6 & 1.4 & 3.0 & 2.0 & 42.9 & 4.2 & 27.6 & 7.5 & 12.8 & 3.0 & 10.9 & 4.8 \\
3DGenZ~\cite{michele2021generative} & $C^S$ & $C^S \cup \hat{C}^U$ & 53.1 & 7.3 & 39.0 & 12.9 & 41.4 & 10.8 & 35.0 & 17.1 & 32.8 & 7.7 & 27.8 & 12.5 \\
SV-Seg~\cite{yang2023zero} & $C^S$ & $C^S \cup \hat{C}^U$ & 58.9 & 9.7 & 43.8 & 16.7 & 46.4 & 12.8 & 39.4 & 20.1 & \underline{34.5} & 14.3 & \underline{30.4} & 20.2 \\


3D-PointZshotS (Ours) & $C^S$ & $C^S \cup \hat{C}^U$ & \underline{68.3} & \underline{12.8} & \underline{51.2} & \underline{21.5} & \underline{54.2} & \underline{14.0} & \underline{45.7} & \underline{22.2} & 34.0 & \underline{14.9} & 30.2 & \underline{20.7} \\

\hline
\end{tabular}
}
\caption{Performance comparison across different methods for generalized zero-shot semantic segmentation on three datasets. $\hat{C}^U$ stands for generated unseen class data. The best values among the generalized zero-shot-learning methods are \underline{underlined}.}
\vspace{-0.2cm}

\label{exp:tab1}
\end{table*}

%
%
%

\subsection{Experimental Setup}
\vspace{-0.1cm}
We adopt the same backbones, generator, and semantic embeddings as the existing baselines~\cite{yang2023zero,michele2021generative} to ensure a commensurate comparison.~Specifically, as backbones, we use ConvPoint~\cite{boulch2020convpoint} for S3DIS, FKAConv~\cite{boulch2020fkaconv} for ScanNet, and  KPConv~\cite{thomas2019kpconv} for SemanticKITTI. Each backbone is pre-trained on seen class data under the best settings proposed in its respective paper.~We employ a generative moment-matching network (GMMN)~\cite{li2015generative} for visual feature generation, and concatenate Word2Vec~\cite{mikolov2013distributed} and GloVe~\cite{pennington2014glove} as 600-dimensional semantic embeddings. We use the Adam~\cite{kingma2014adam} optimizer with an initial learning rate of 0.0002 for ScanNet, 0.0001 for S3DIS, and 0.007 for SemanticKITTI. We set $\tau_1$ to 0.5 and $\tau_2$ to 0.2. $M$, which is the number of LGPs, is set to 128.

\subsection{Datasets}
\vspace{-0.1cm}
We conduct experiments on three datasets, including two indoor datasets, S3DIS~\cite{armeni2017joint} and ScanNet~\cite{dai2017scannet}, and one outdoor dataset, SemanticKITTI~\cite{behley2019semantickitti}. Following~\cite{yang2023zero,michele2021generative}, we consider four unseen classes for each dataset, and the remaining classes are set as seen classes.
S3DIS is a comprehensive point cloud dataset widely used for 3D indoor scene understanding.~It consists of 3D scans of 6 large indoor areas, including 271 scanned rooms covering 13 classes. Area 5 is held out for testing. We set ``beam'', ``column'', ``window'' and ``sofa'' as unseen classes.
ScanNet is another large-scale dataset. It includes over 1,500 indoor scans with 20 classes, primarily collected from apartments and office environments. We set ``desk'', ``bookshelf'', ``sofa'' and ``toilet'' as unseen classes. The standard validation set is used for testing.
SemanticKITTI is a large-scale dataset containing 21 sequences of street-scape point cloud data with 19  classes. Sequence 08 is used for testing. The unseen classes are ``motorcycle'', ``truck'', ``bicyclist'' and ``traffic-sign''. 


\subsection{Comparison with Baselines}
\begin{figure*}[h]
  \centering
   \includegraphics[width=\linewidth]{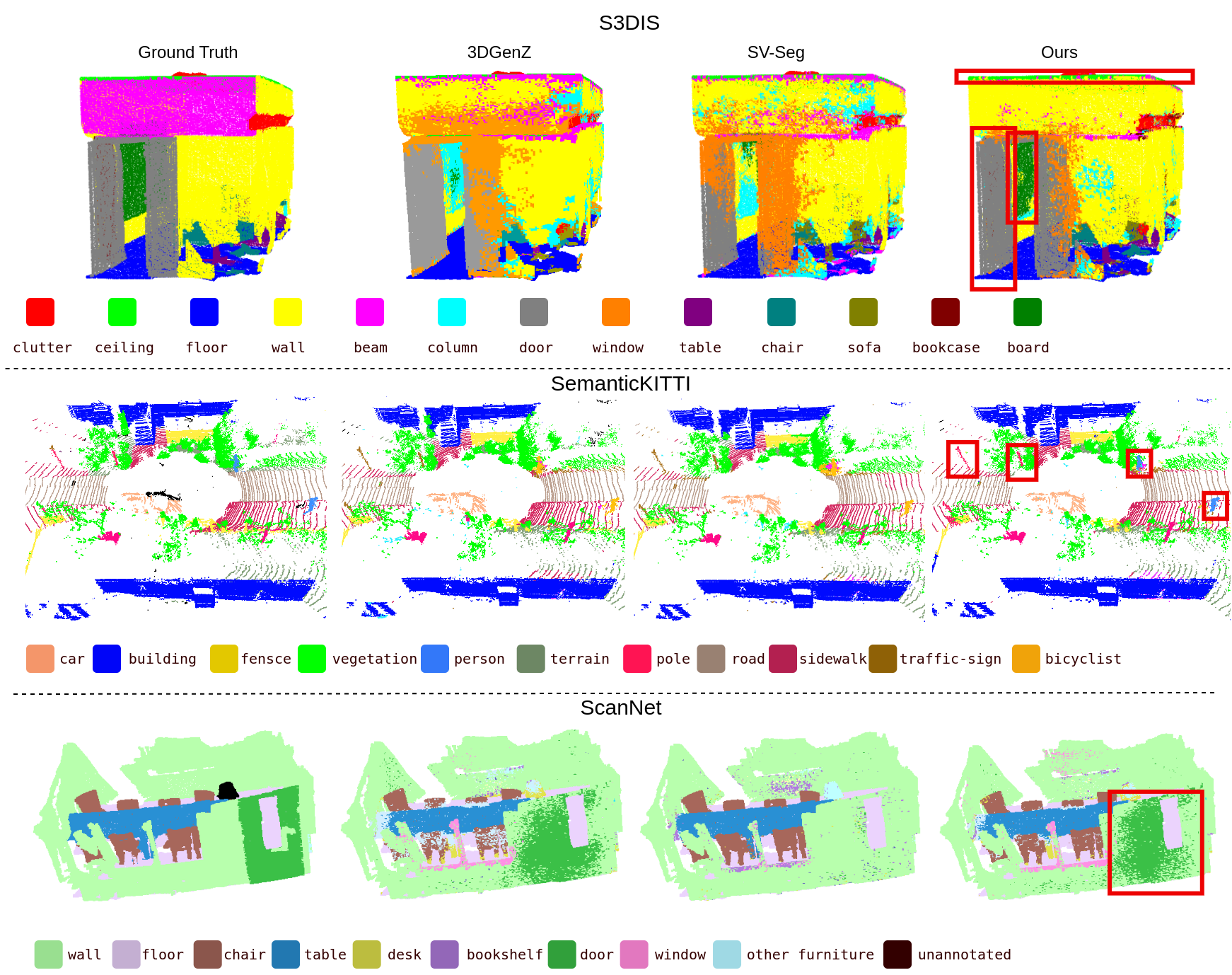}
   \caption{Qualitative comparison on three datasets: the first column shows the ground truth; the second, 3DGenZ predictions; the third, SV-Seg predictions; and the fourth, our predictions. Red rectangles indicate regions where our method performs better. Best viewed when zoomed in.}
   \vspace{-0.4cm}
   \label{fig:qualitative}
\end{figure*}

\vspace{-0.1cm}
We use mean Intersection-over-Union (mIoU) and Harmonic mIoU (HmIoU) as evaluation metrics. Since the results may be biased toward seen classes, HmIoU can assess the balance in performance between seen and unseen classes and provide a more holistic view of model performance. Tab.~\ref{exp:tab1} presents the results across three datasets, providing a comparison with two baseline zero-shot point cloud semantic segmentation methods (3DGenZ~\cite{michele2021generative} and SV-Seg~\cite{yang2023zero}), two zero-shot learning (ZSL) baselines adapted for 3D zero-shot segmentation, and three supervised methods. The ZSL baselines include ZSLPC-Seg, which adapts the ZSL method ZSLPC~\cite{cheraghian2019zero} for segmentation, and DeViSe-3DSeg, an adaptation of DeViSe~\cite{frome2013devise} from image classification to point cloud segmentation. Moreover, following the setup in~\cite{michele2021generative}, we include three supervised models for reference: (1) Full supervision, where the entire model (backbone and classifier) is trained on a fully annotated dataset of seen and unseen classes; (2) ZSL backbone, in which the backbone is trained on seen classes and the classifier on seen and unseen classes; and (3) ZSL-trivial, where the entire model is trained only on seen classes. All baseline results are taken from~\cite{michele2021generative}. We also visualize the segmentation results of two baselines and our method on three datasets in Fig.~\ref{fig:qualitative}.

%
%

In terms of unseen class mIoU, our method outperforms all baselines, demonstrating its effectiveness. For S3DIS, our method significantly surpasses SV-Seg by 10.6\% and 3.1\% in mIoU on seen and unseen classes, respectively, and achieves a 4.8\% improvement in HmIoU. The first row in Fig.~\ref{fig:qualitative} shows that our method achieves better performance for some classes, like ``door'' and ``board''. These classes have well-defined geometric attributes. The incorporation of LGPs likely enhances feature discrimination by embedding this characteristic structures, leading to better segmentation performance. For classes, ``beam'' and ``column'', however, our method provides lower IoU, which likely arises from the fact that their semantic and/or visual embeddings may lack sufficient distinctiveness.

For SemanticKITTI, our method not only achieves better seen and unseen class mIoU than SV-Seg but also surpasses the ZSL supervised backbone.~Compared to SV-Seg, our method provides substantially higher IoU for classes, such as ``person'' and ``pole'', as shown in Fig.~\ref{fig:qualitative}. The ZSL supervised backbone setup helps assess how well the representation learned from seen classes can generalize to unseen classes through a linear classifier.~In contrast, our method re-represents the visual and semantic features in a geometry-based common space and employs a nearest-neighbor-based classifier.~This adjustment enhances the model's transferability and generalization across classes. 

For ScanNet, our method achieves better unseen class mIoU and HmIoU compared to SV-Seg, though the improvement margins are smaller. This is likely because ScanNet 
contains a higher number of semantically similar classes. The increased similarity between classes makes it more difficult for the classifier to distinguish between them, thereby reducing the overall improvement margin.


In the supplementary material, we include ablation studies validating LGP integration, cohesive visual-semantic representations, and the self-consistency loss, along with analyses on the number of LGPs and semantic embeddings.

%% file: sec/5_conclusion.tex
\section{Conclusion}
\label{sec:conclusion}

In this paper, we have incorporated latent geometric prototypes into both feature generation and alignment. These latent geometric prototypes capture geometric attributes of point cloud data and are shareable across seen and unseen classes. We have proposed a geometric consistency-aware generator to synthesize visual features enriched with fine-grained geometric attributes, using an InfoNCE-based self-consistency loss to ensure feature stability under random point sampling. To bridge the semantic-visual domain gap, we re-represent the visual and semantic features based on their similarity with latent geometric prototypes, creating a cohesive visual-semantic representation that facilitates the knowledge transfer from semantics to visual data. Our method achieves state-of-the-art performance on three scene-level point cloud datasets in terms of HmIoU.

%% file: sec/suppl.tex
\clearpage
\setcounter{page}{1}
\maketitlesupplementary

In this supplementary material, we first examine the necessity and role of incorporating latent geometric prototypes (LGPs) for feature generation and alignment, as well as the impact of the cohesive visual-semantic representation and self-consistency loss. Next, we present the experiments evaluating the effects of varying the number of LGPs and selecting different semantic embeddings. \vspace{0.2cm}

\noindent \textbf{1. Effect of integrating LGPs in the generator}

To demonstrate the effectiveness of using LGPs, we only use the semantic features and the random noise as input to the generator.~As shown in Tab.~\ref{tab:abs2}, although the model without the geometry-aware generator (1st row) can achieve similar seen class mIoU, its unseen class mIoU and HmIoU are much lower than the full model (4rd row). We believe that feeding the geometry-aware semantic features to the generator facilitates addressing the seen and unseen class bias issue and bridges the domain gap.
\begin{table}[h!]
\vspace{-0.1cm}
\centering
\resizebox{\linewidth}{!}{
\begin{tabular}{c|c|c|ccc|c}
\hline
\multirow{2}{*}{Generator w/ LGPs} & \multirow{2}{*}{Generator w/ $\mathcal{L}^{self}$} & \multirow{2}{*}{Align Vis-Sem.} & \multicolumn{3}{c|}{mIoU}       & \multirow{2}{*}{HmIoU} \\ \cline{4-6}
                           &              &                & $C^s$                        & $C^u$                        & All  &                        \\ \hline
     \ding{55}                &  \ding{51}            &               \ding{51}               & \multicolumn{1}{c|}{68.0} & \multicolumn{1}{c|}{8.5}  & 49.6 & 15.0                   \\ \hline
      \ding{51}               &  \ding{51}          &          \ding{55}                    & \multicolumn{1}{c|}{40.4} & \multicolumn{1}{c|}{6.7}  & 30.1 & 11.6                   \\ \hline
      \ding{51}    & \ding{55}   & \ding{51} &         \multicolumn{1}{c|}{67.7} & \multicolumn{1}{c|}{10.5} & 50.1 & 18.2                  \\ \hline
         \ding{51}             &  \ding{51}          &          \ding{51}                     & \multicolumn{1}{c|}{68.3} & \multicolumn{1}{c|}{12.8} & 51.2 & 21.5                   \\ \hline
\end{tabular}
} \vspace{0.15cm}
\caption{Ablation study of the effects of (i) using LGPs in the generator and (ii) aligning the visual and semantic features in a shared space.  
}
\label{tab:abs2}
\end{table}

\noindent  \textbf{2. Effect of cohesive visual-semantic representation}

To demonstrate the necessity of transforming semantic features to the latent geometric space, we conduct another experiment, where semantic features are projected into the same dimensional space as the visual features using a linear layer. As shown in the second row of Tab.~\ref{tab:abs2}, this approach led to a significant mIoU drop of 27.9\% on seen classes and a drop of 6.1\% on unseen classes compared to the full model. Since point-level classification is based on the similarity between the visual and semantic features, a simple projection of semantic features might not be sufficient for effective feature alignment.\vspace{0.2cm}

\noindent  \textbf{3. Effect of adding the self-consistency loss}

We remove the loss term $\mathcal{L}^{self}$, which is designed to improve the self-consistency of synthetic visual features. As indicated in the third row of Tab.~\ref{tab:abs2}, the seen mIoU drops to $67.7\%$, while the unseen class mIoU drops to $10.5\%$, showing that the self-consistency loss helps generate robust and consistent representations for unseen classes, while seen classes remain less affected due to the availability of visual data. \vspace{0.2cm}

\noindent \textbf{4. Visualization of Latent Geometric Prototypes}

\begin{figure}[b!]
  \centering
   \includegraphics[width=\linewidth]{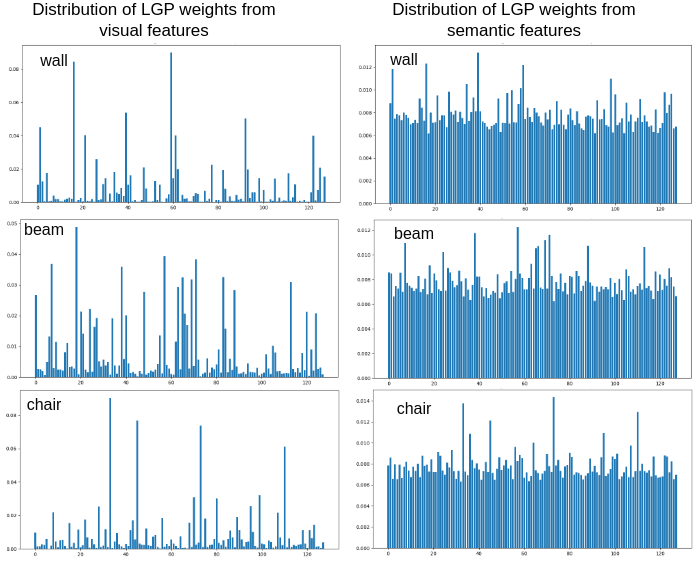}
   \caption{The distribution of LGP weights from visual and semantic features.}
   \label{fig:gp_dist}
   \vspace{-0.3cm}
\end{figure}

Since the LGPs are shaped across semantic and visual features, we aim to achieve a similar frequency distribution for the semantic and visual features representing the same class. In Fig.~\ref{fig:gp_dist}, we visualize the distribution of LGPs for both visual and semantic features, showing a noticeable difference for each class. Semantic feature distributions tend to be relatively uniform, while visual feature distributions are more concentrated.
This difference arises because word embeddings provide abstract and generalized representations that spread uniformly across multiple LGPs. In contrast, visual features derived from 3D data capture high-level geometric features, aligning strongly with particular LGPs and creating a concentrated distribution. 

Despite these differences, re-representation aligns semantic and visual features within a shared space.~We also observe that visually similar classes, like ``beam'' and ``wall'', exhibit closer LGP distributions than dissimilar classes, such as ``chair''. This alignment of LGP distributions for similar classes can support ZSL by enhancing transferability and increasing robustness across similar classes.
\vspace{0.2cm}

\noindent \textbf{5. Effect of the number of LGPs}

We conduct more experiments to evaluate the effect of using different numbers of LGPs. As shown in Tab.~\ref{tab:num_gps}, the performance, in terms of both metrics, improves for both seen and unseen classes as the number of LGPs increases from 32 to 128, with 128 providing the highest performance in terms of HmIoU,  unseen mIoU and all mIoU. However, increasing the number of LGPs to 192 results in a decline in unseen mIoU and HmIoU, while increasing the mIoU for seen classes. This suggests that too many LGPs may lead to overfitting to seen classes or introduce redundancy, reducing the model's ability to generalize.

\begin{table}[h!]
\centering
\resizebox{0.45\textwidth}{!}{
\begin{tabular}{c|c|c|c|c}
\hline
      & seen mIoU & unseen mIoU & all mIoU & HmIoU \\ \hline
M=32  & 67.3      & 8.7         & 49.3     & 15.5  \\ \hline
M=64  & 67.7      & 12          & 50.5     & 20.4  \\ \hline
M=128 & 68.3      & \textbf{12.8}        & \textbf{51.2}     & \textbf{21.5}  \\ \hline
M=192 & \textbf{68.7}      & 11.7        & \textbf{51.2}     & 20    \\ \hline
\end{tabular}
}
\vspace{0.2cm}
\caption{Effect of using different number ($M$) of LGPs. Using M=128 achieves the optimal balance between mIoUs for seen and unseen classes.}
\label{tab:num_gps}
\end{table}

\noindent \textbf{6. Effect of semantic embeddings}

To verify the effectiveness of semantic features, we conduct experiments using Word2Vec~\cite{mikolov2013distributed} and GloVe~\cite{pennington2014glove} as 300-dimensional features by themselves, as well as concatenating them. As shown in Tab.~\ref{tab:sem_embd}, the mIoU for unseen classes decreases when using either GloVe or Word2Vec alone, with a significant drop observed for Word2Vec. This can be attributed to the differences in their pre-training methodologies.~GloVe captures global co-occurrence statistics, which provide a more holistic representation of word relationships. Word2Vec, on the other hand, focuses on local context through a sliding window. When re-representing semantic features via LGPs, GloVe might align better with the geometric attributes than Word2Vec. However, combining GloVe and Word2Vec achieves the highest mIoU for unseen classes, leveraging the strengths of both embeddings.

\begin{table}[hb!]
\centering
\resizebox{0.45\textwidth}{!}{
\begin{tabular}{c|c|c|c|c}
\hline
         & seen mIoU & unseen mIoU & all mIoU & HmIoU \\ \hline
GloVe    & 68.2      & 10.9        & 50.6     & 18.8  \\ \hline
Word2Vec & 68.2      & 5.7         & 48.9     & 10.6  \\ \hline
GloVe+Word2Vec                   & \textbf{68.3}     & \textbf{12.8}        & \textbf{51.2}     & \textbf{21.5}  \\ \hline
\end{tabular}
}
\vspace{0.2cm}
\caption{Effect of using different semantic embeddings. Combining GloVe and Word2Vec achieves the best HmIoU as well as best overall mIoU.}
\label{tab:sem_embd}
\end{table}